\pdfoutput=1

\documentclass[11pt]{article}

\usepackage[final]{acl}

\usepackage{times}
\usepackage{latexsym}

\usepackage[T1]{fontenc}

\usepackage[utf8]{inputenc}

\usepackage{microtype}

\usepackage{inconsolata}

\usepackage{graphicx}

\usepackage{amsmath}
\usepackage{booktabs, multirow}
\usepackage{listings}

\usepackage{tikz}
\def\checkmark{\tikz\fill[scale=0.4](0,.35) -- (.25,0) -- (1,.7) -- (.25,.15) -- cycle;} 

\title{SignCLIP: Connecting Text and Sign Language by Contrastive Learning}

\author{
\parbox{0.6\linewidth}{\centering
Zifan Jiang, Gerard Sant, Amit Moryossef, \\
Mathias Müller, Rico Sennrich, Sarah Ebling}
\\
University of Zurich
\\
\texttt{jiang@cl.uzh.ch}}

\begin{document}
\maketitle
\begin{abstract}

We present SignCLIP, which re-purposes CLIP (Contrastive Language-Image Pretraining) to project spoken language text and sign language videos, two classes of natural languages of distinct modalities, into the same space.
SignCLIP is an efficient method of learning useful visual representations for sign language processing from large-scale, multilingual video-text pairs, without directly optimizing for a specific task or sign language which is often of limited size.

We pretrain SignCLIP on \textit{Spreadthesign}, a prominent sign language dictionary consisting of $\sim$500 thousand video clips in up to 44 sign languages, and evaluate it with various downstream datasets.
SignCLIP discerns in-domain signing with notable text-to-video/video-to-text retrieval accuracy.
It also performs competitively for out-of-domain downstream tasks such as isolated sign language recognition upon essential few-shot prompting or fine-tuning. 

We analyze the latent space formed by the spoken language text and sign language poses, which provides additional linguistic insights.
Our code and models are openly available\footnote{\url{https://github.com/J22Melody/fairseq/tree/main/examples/MMPT}}.
\end{abstract}

\section{Introduction}
\label{sec:intro}

Sign(ed) languages are the primary communication means for $\sim$70 million deaf people worldwide\footnote{
\url{https://wfdeaf.org/our-work/}}.
They use the visual-gestural modality to convey meaning through manual articulations in combination with non-manual elements like the face and body \citep{sandler2006sign}. 
Sign language processing (SLP) \citep{bragg2019sign, yin-etal-2021-including} is a subfield of natural language processing (NLP) that is intertwined with computer vision (CV) and sign language linguistics.

\begin{figure}
    \centering
    \includegraphics[width=\linewidth]{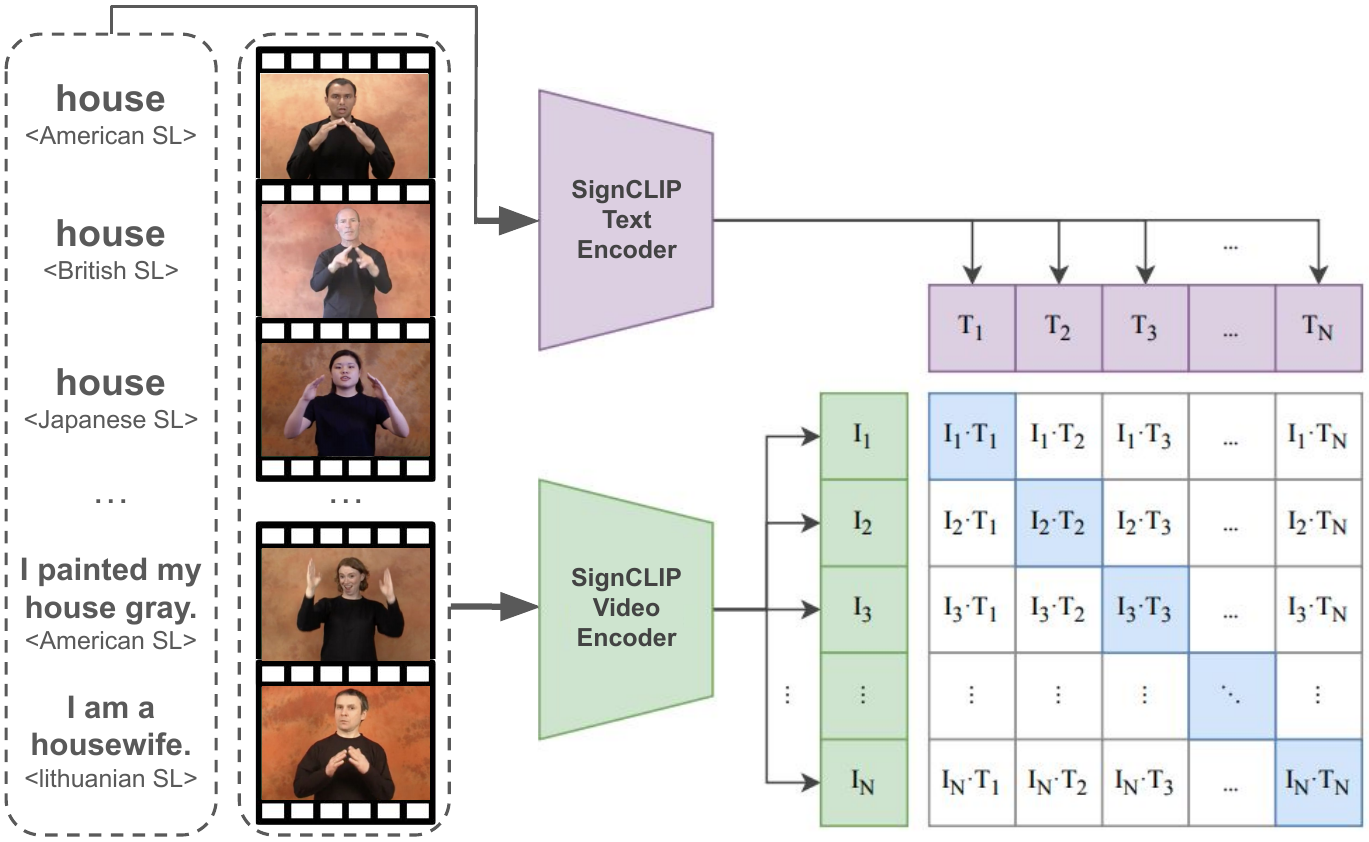}
    \caption{Illustration of SignCLIP, comprising a text encoder and a video encoder jointly trained on pairs of text and multilingual signing examples. Every sign is articulated in diverse languages and contexts with subtle differences in hand shape, movement, place of articulation, etc. The screenshots of the videos are from \textit{Spreadthesign} and the matrix part is taken from CLIP.}
    \label{fig:title}
\end{figure}

SLP spans the tasks of sign language recognition \citep{adaloglou2021comprehensive}, translation \citep{de2023machine}, and production \citep{rastgoo2021sign}. 
Typical datasets, equipped with spoken language text and sign language glosses in addition to videos, that support SLP research are RWTH-PHOENIX-Weather 2014T, in German Sign Language (DGS), introduced by \citet{forster-etal-2014-extensions, camgoz2018neural}; and CSL-Daily, in Chinese Sign Language (CSL), introduced by \citet{zhou2021improving}.
However, advances on a specific task/dataset/language are often limited and non-transferable to more generic and challenging settings \citep{muller-etal-2022-findings, muller-etal-2023-findings} due to the small, domain-specific vocabulary (1,066 and 2,000 signs, respectively) and data size (11 and 23 hours, respectively).
Recent sign language corpora with a scale of more than a thousand signing hours have emerged for relatively high-resource sign languages, e.g., BOBSL \citep{albanie2021bobsl} for British Sign Language (BSL) and YouTube-ASL \citep{uthus2024youtube} for American Sign Language (ASL), a subset of YouTube-SL-25 \citep{tanzer2024youtube}. JWSign \citep{gueuwou-etal-2023-jwsign} is also a notable corpus that consists of 2,530 hours of Bible translations in 98 sign languages.

In the meanwhile, outside the world of SLP, there is great progress in deep pretrained models of different modalities, GPT \citep{achiam2023gpt} for text, masked autoencoders \citep{he2022masked} for images, and wav2vec 2.0 \citep{baevski2020wav2vec} for speech, to name a few.
They are commonly pretrained on a huge amount of data (e.g., over 15 trillion tokens for Llama 3, \citet{touvron2023llama}) with very little or weak supervision, but present striking multilingual and multi-task ability, for zero-shot prediction, fine-tuning, and representation learning. 

Ideally, the visual aspect of sign languages, i.e., lexical similarity due to iconicity \citep{johnston-schembri-2007} (illustrated in Figure \ref{fig:title}) 
makes transferring between different sign languages easier than the text of different spoken languages or even scripts. The latter faces unfair tokenization issues \citep{petrov2024language} and out-of-vocabulary errors. 
At the same time, unlike discrete text tokens (see the comparison in Table \ref{tab:encoders-dimensions}), the dense continuous video signal is expensive to process computationally and seems daunting for the above-mentioned self-supervised training approaches.

Since SLP tasks and datasets usually involve both the text and visual/signed modalities, we take inspiration from OpenAI's CLIP model \citep{radford2021learning} but use contrastive learning to connect text with sign language videos instead of images.
For video understanding, follow-up work such as VideoCLIP \citep{xu-etal-2021-videoclip} mainly deals with tasks including action recognition \citep{zhu2020comprehensive} and VideoQA \citep{xu2016msr, yu2018joint}.
However, both CLIP, VideoCLIP, and other existing multimodal models understand visual content on a coarse-grained level and generic domain and do not address the intricacy of sign language.
We show the lack of sign language understanding in contemporary AI models both intuitively (Figure \ref{fig:house} in Appendix \ref{appendix:chatgpt}) and empirically (Table \ref{tab:finger_clip}).
 
This work uses sign-language-specific data to train a CLIP-like model for SLP.
We first validate the approach's feasibility on fingerspelling, a subsystem of sign language, by a model named FingerCLIP (\S\ref{sec:finger_clip}), which correctly understands the fingerspelling of individual letters.
We then curate the Spreadthesign\footnote{\url{https://www.spreadthesign.com/}. The use of the data is under a license granted by Spreadthesign.} 
dictionary as a large-scale pretraining dataset consisting of $\sim$500 hours of signing, as well as diverse public downstream task datasets 
to run comprehensive pretraining, fine-tuning, and evaluation on full-fledged sign languages.
By contrastive training on $\sim$500 thousand video-text pairs, we obtain a multimodal and multilingual model named SignCLIP (\S\ref{sec:sign_clip}), illustrated by Figure \ref{fig:title}. SignCLIP excels at various SLP tasks and datasets and presents a compelling latent space for signed video content aligned with spoken language text (\S\ref{sec:evaluation}). 

\section{Background: Sign Language Representation}
\label{sec:background}

Representation is a key challenge for SLP. 
Unlike spoken languages, sign languages have no widely adopted written form. As sign languages are conveyed through the visual-gestural modality, video recording is the most straightforward way to capture them. 
The final goal of SignCLIP is to represent a sign language video clip by an embedding that aligns with a ubiquitous text encoder like Sentence-BERT \citep{reimers-2019-sentence-bert}.
Generally, end-to-end training on the raw videos is computationally costly, and various intermediate representations alleviate this issue.

\paragraph{VideoCLIP and Video Encoders}
Our work adapts VideoCLIP, which is pretrained by general instructional videos from the HowTo100M \citep{miech19howto100m} dataset.
We aim at replacing HowTo100M with domain-specific sign language videos, such as the dataset How2Sign \citep{Duarte_CVPR2021}, albeit on a considerably smaller scale.

Videos are very dense temporally (frame rate) and spatially (video resolution).
A 3D-CNN-based video encoder is often used to extract informative features with reduced dimensionalities for downstream tasks.
VideoCLIP uses an S3D \citep{zhang2018s3d} model pretrained on HowTo100M that produces one video token (i.e., a video embedding for the temporal window) per second. 
For SLP, it is possible to use a video encoder pretrained specifically on sign language videos. A prominent one is the I3D \citep{carreira2017quo} model pretrained on the BSL sign language recognition task \citep{Varol21} with the BSK-1K dataset \citep{Albanie2020bsl1k}. 
A more recent approach to simultaneously address temporal and spatial complexity is the Video Swin Transformer proposed by \citet{liu2022video}, and \citet{Prajwal22a} trains one such model for BSL fingerspelling recognition.

\paragraph{Pose Estimation}
A potentially more interpretable and universal way of extracting sign language-related features from videos is human pose estimation \citep{zheng2020deep}, for example, using MediaPipe Holistic \citep{lugaresi2019mediapipe, mediapipe2020holistic}. 
Each video frame is converted into the location (X, Y, Z) of 543 full-body keypoints in a 3D space.
However, the immediate applicability of the pose estimation systems for SLP is questionable \citep{moryossef2021evaluating}, and known issues such as the lack of accurate depth information are presented \citep{holmes2024key}. 
Generally, there is a trade-off between user-friendliness and accuracy for pose estimation tools and SLP researchers often prefer MediaPipe Holistic for the former \citep{selvaraj-etal-2022-openhands} over others like OpenPose \citep{openpose} and AlphaPose \citep{alphapose}.
An even more universal alternative approach to track keypoints named \textit{Tracking Everything Everywhere All at Once} is proposed by \citet{wang2023omnimotion}. \citet{sevilla-etal-2024-automated-extraction} uses a similar CoTracker \citep{karaev2023cotracker} model to study sign language prosody. 
Such models produce smoother signals than traditional pose estimation.

\paragraph{Discrete Representation}
Non-standard written forms of sign language, including SignWriting \citep{writing:sutton1990lessons}, HamNoSys \citep{writing:prillwitz1990hamburg}, and glosses \citep{Johnston2008FromAT}, offer the possibility to incorporate sign language content into a text-based NLP pipeline \citep{jiang-etal-2023-machine}.
However, a good segmentation \citep{moryossef-etal-2023-linguistically} and transcription model to process raw video input is first required, which is not well-researched and is not part of this paper.

Recently, vector quantization (VQ) approaches \citep{van2017neural} such as SignVQNet \citep{hwang2023autoregressive} demonstrate the ability to convert the continuous signal of videos/poses to discrete tokens similar to spoken language sub-words \citep{sennrich-etal-2016-neural}, which might be a promising direction to pursue in future work. 

\paragraph{Comparison}
In this work, we only empirically experiment with the video encoder and pose-based methods since there are not yet mature and open solutions for the others at the time of writing.
Given a hypothetical 10-second, 30 FPS, 480p (640×480), RGB (3 channels) video of 12 consecutive signs, we compare the dimensionalities of the most common representations in Table \ref{tab:encoders-dimensions}. 
These approaches compress 
the raw videos to a sequence of video tokens compatible with a Transformer \citep{vaswani2017attention} or a pretrained language model for further training and processing \citep{gong2024llms}.

\begin{table}
    \centering
    \resizebox{\linewidth}{!}{%
        \begin{tabular}{l|lll}
        \toprule
        
        \textbf{Representation} & \textbf{Temporal} & \textbf{Spatial} & \textbf{Interpretable} \\ 
        \midrule
        
        Original video & 10x30 & 640$\times$480$\times$3 & -- \\ \midrule
        S3D (HowTo100M) & 10 & 512 & no \\ \midrule
        I3D (BSL-1K) & 10 & 1024 & no \\ \midrule
        MediaPipe Holistic & 10$\times$30 & 543$\times$3 & yes \\ \midrule
        VQ (e.g., SignVQNet) & 10 & 1024* & no \\ \midrule
        SignWriting/HamNoSys/gloss & 12 & 1024* & yes \\ 
        \bottomrule
        
        \end{tabular}
    }
    \caption{Temporal and spatial dimensions of different sign language representations for a 10-second, 30 FPS, 480p (640×480), RGB (3 channels) video consisting of 12 signs. In the parentheses are the datasets on which the models are pretrained. For discrete representations, we assume the embedding size to be 1024, marked with an asterisk (*), but this can be chosen arbitrarily.}
    \label{tab:encoders-dimensions}
\end{table}


\section{Model Architecture}

We follow the setups in VideoCLIP and reuse their codebase\footnote{\url{https://github.com/facebookresearch/fairseq/tree/main/examples/MMPT}}. 
The most essential model architecture with minor modifications to adapt our experiments is described here.
We take pairs of video and text samples $(v, t)$ as inputs, where for each video clip, $\boldsymbol{c_v}$ is a sequence of continuous video frames and is processed by a video encoder $f_{\theta_{ve}}$. 
This is then followed by a trainable MLP projection layer, $f_{\theta_{vp}}$, to project the embedding to the same dimension, $d=768$, as the word embedding on the text side:

\begin{equation}
\boldsymbol{x_v} = f_{\theta_{vp}}(stopgrad(f_{\theta_{ve}}(\boldsymbol{c_v}))
\end{equation}

The frozen video encoder $f_{\theta_{ve}}$ is by default a 3D-CNN network but can be replaced by any other visual backbone or a black-box pose estimation system as summarized in Table \ref{tab:encoders-dimensions}. 
Likewise, text token vectors $\boldsymbol{x_t}$ are acquired through embedding lookup from a frozen BERT model \citep{devlin-etal-2019-bert}. 
Then $\boldsymbol{x_v}$ and $\boldsymbol{x_t}$ are fed into two separate
trainable Transformers, $f_{\theta_{v}}$ and $f_{\theta_{t}}$, followed by average pooling over the sequence of the tokens to obtain the temporally aggregated embeddings:

\begin{equation}
z_v = Avg(f_{\theta_{v}(\boldsymbol{x_v})}), 
z_t = Avg(f_{\theta_{t}(\boldsymbol{x_t})})
\end{equation}

We optionally add two linear multimodal projection layers on top of $\boldsymbol{z_v}$ and $\boldsymbol{z_t}$, which are missing in VideoCLIP but present in CLIP (see Figure 3 of the CLIP paper).
Finally, we employ the InfoNCE loss \citep{oord2018representation} as the contrastive objective to discern the relationship between the embedded $N$ video-text pairs in each mini-batch and run contrastive training over the whole dataset.

\section{FingerCLIP}
\label{sec:finger_clip}

As a proof of concept, we first apply this contrastive training approach to
Fingerspelling \citep{battison1978lexical,wilcox1992phonetics,brentari2001language}, a subsystem of sign languages heavily influenced by the surrounding spoken languages. 
For concepts that do not (yet) have associated signs (names of people, locations, organizations, etc.), sign language users borrow a word of a spoken language by spelling it letter-by-letter with predefined signs for that language's alphabet. 
Isolated fingerspelling recognition\footnote{Continuous fingerspelling recognition is more complex and is often solved by a CTC loss like speech recognition. For more thorough research on fingerspelling, we recommend the ChicagoFSWild+ dataset and Google's ASL fingerspelling recognition competition on Kaggle.} can therefore be considered a toy task similar to the MNIST \citep{deng2012mnist} handwritten digits classification task in CV.
We name the model FingerCLIP, a mini-version of SignCLIP (\S\ref{sec:sign_clip}).


\paragraph{Dataset}
We start with the RWTH German Fingerspelling Database \citep{dreuw2006modeling}, containing $\sim$1400 videos of 35 DGS fingerspelling and number signs, five of which contain inherent motion. 
We provide details and an illustration of the dataset in Appendix \ref{appendix:rwthfs}. 
We split all examples randomly into training/validation/test sets at the ratio of 8:1:1.

\begin{table*}[htbp]

\centering
\resizebox{\textwidth}{!}{%
\begin{tabular}{ll|cccc|cccc}

\toprule

 


 & & \multicolumn{4}{c}{\textbf{Text-to-video}} & \multicolumn{4}{c}{\textbf{Video-to-text}} \\
 
\cmidrule(lr){3-6} \cmidrule(lr){7-10}

\multicolumn{2}{l}{\textbf{Experiment}} & \textbf{P@1}$\uparrow$ & \textbf{P@5}$\uparrow$ & \textbf{P@10}$\uparrow$ & \textbf{MedianR}$\downarrow$ & \textbf{R@1}$\uparrow$ & \textbf{R@5}$\uparrow$ & \textbf{R@10}$\uparrow$ & \textbf{MedianR}$\downarrow$ \\
\midrule

\textbf{E0} & \citet{dreuw2006modeling} (supervised HMM, appearance-based features) & -- & -- & -- & -- & 0.64 & -- & -- & -- \\
\midrule

\multicolumn{2}{l}{\textit{Explore training strategy}} & \\
\textbf{E1} & VideoCLIP zero-shot (+ \textit{S3D HowTo100M} video features) & 0.03 & 0.02 & 0.03 & 22 & 0.02 & 0.14 & 0.28 & 18 \\
\textbf{E1.1} & VideoCLIP fine-tuned (+ \textit{S3D HowTo100M} video features) & 0.40 & 0.36 & 0.30 & 2 & 0.31 & 0.75 & 0.89 & 2 \\
\textbf{E1.2} & VideoCLIP trained from scratch (+ \textit{S3D HowTo100M} video features) & 0.54 & 0.35 & 0.28 & \textbf{1} & 0.28 & 0.69 & 0.87 & 3 \\
\midrule

\multicolumn{2}{l}{\textit{Explore video-based (I3D) features}} & \\
\textbf{E2} & FingerCLIP trained from scratch (+ \textit{I3D BSL-1K} video features) & 0.63 & 0.47 & 0.37 & \textbf{1} & 0.37 & 0.78 & 0.91 & 2 \\
\textbf{E2.1} & \textbf{E2} + feature dimension average pooled from 1024 to 512 & 0.74 & 0.56 & \textbf{0.44} & \textbf{1} & 0.47 & 0.82 & 0.94 & 2 \\
\midrule

\multicolumn{2}{l}{\textit{Explore pose-based features}} & \\
\textbf{E3} & FingerCLIP trained from scratch (\textit{MediaPipe Holistic} pose features) & 0.89 & 0.67 & 0.42 & \textbf{1} & 0.68 & 0.97 & \textbf{1.00} & \textbf{1} \\
\textbf{E3.1} & \textbf{E3} + dominant hand features only (26 times less keypoints) & \textbf{1.00} & 0.72 & 0.42 & \textbf{1} & 0.82 & 0.99 & \textbf{1.00} & \textbf{1} \\
\textbf{E3.2} & \textbf{E3.1} + 2D augmentation on pose features ($\sigma=0.2$) & 0.91 & \textbf{0.74} & 0.43 & \textbf{1} & \textbf{0.93} & \textbf{1.00} & \textbf{1.00} & \textbf{1} \\

\bottomrule
\end{tabular}
}
\caption{FingerCLIP experimental results evaluated on the test set. \textit{P@k} denotes \textit{precision@k}, \textit{R@k} denotes \textit{recall@k}, and \textit{MedianR} denotes the median retrieval rank. The best score of each column is in bold. \textit{E0} is taken from \citet{dreuw2006modeling} as a baseline (\textit{R@1} derived from the best error rate 35.7\%).}
\label{tab:finger_clip}
\end{table*}

\paragraph{Training Details}
We adhere to most implementation details outlined in VideoCLIP unless otherwise specified. 
The two trainable Transformers, $f_{\theta_{v}}$ and $f_{\theta_{t}}$, are initialized with the pretrained \textit{bert-base-uncased} weights. 
We train 25 epochs within two hours on a Tesla V100-SXM2-32GB GPU, validated by loss on the validation set. 
For contrastive training, we construct each batch as a collection of 35 different signs with corresponding text prompts ``Fingerspell the letter <letter\_name> in DGS.''. 
By optimizing the InfoNCE loss, we move the embedding of a sign closer to the paired text, and further away from the remaining 34 negative examples.

We test different combinations of video encoders:
(a) \textit{S3D HowTo100M} video features\footnote{\url{https://github.com/antoine77340/S3D_HowTo100M}};
(b) \textit{I3D BSL-1K (M+D+A)} video features\footnote{\url{https://www.robots.ox.ac.uk/~vgg/research/bslattend/}};
(c) \textit{MediaPipe Holistic} pose estimation,
and training strategies:
(a) zero-shot VideoCLIP (no training);
(b) fine-tuning VideoCLIP;
(c) training from scratch.

MediaPipe Holistic runs offline on a low-end CPU device.
Pose estimation is normalized to a consistent scale by setting the mean width of each person's shoulders to $1$, and the mid-point to $(0,0)$. 
The leg values are removed since they are irrelevant to signing. 
We further augment the data by randomly rotating, shearing, and scaling the poses\footnote{The \href{https://github.com/sign-language-processing/pose}{Pose} library implements related operations.}. 

\paragraph{Evaluation}
We view fingerspelling understanding as a text-to-video/video-to-text retrieval task. The candidates are ranked for both directions by a dot-product-based similarity score to each text/video query in the latent space.
For the test text prompt of each sign, there is possibly more than one correct video (e.g., the same letter signed by different signers) in the test video pool, and they are all considered successful retrieval. 
We thus evaluate the text-to-video retrieval task by \textit{precision@k}, i.e., what percent of the top k candidates are correct answers. 
Each test video query has only one correct text prompt out of the 35 possible prompts. 
We thus evaluate the video-text retrieval task by \textit{recall@k}, i.e., the chance to include the only correct answer by taking the top k candidates. 
\textit{precision@1} and \textit{recall@1} can be interpreted as the retrieval accuracy, and in our video-to-text scenario where there is only one relevant text item, \textit{recall@k} also equals top-n accuracy. We keep using  \textit{recall@k} instead of accuracy in this paper for genericity.
We also add the metric median retrieval rank, i.e., the median value of the rank of the first correct answer in the candidate lists.
We present the experimental results in Table \ref{tab:finger_clip}.

\paragraph{Discussion}

FingerCLIP distinguishes itself from the supervised baseline method \citep{dreuw2006modeling} in that it is not directly optimized for classification. Instead, a contrastive objective ties positive pairs of text and videos by learning meaningful embeddings, which are then used for similarity-based retrieval. 
We find video-to-text retrieval reasonably more challenging than text-to-video retrieval\footnote{Note that the relatively low \textit{precision@5/10} values are uncomparable to the high \textit{recall@5/10} values, since the former makes the task harder, while the latter simplifies the task.} since text-to-video retrieval is also often of less value, as a trivial dictionary look-up does the job perfectly.

\textit{E1}, zero-shot VideoCLIP, presenting random guess results, shows that a common video understanding network pretrained on HowTo100M does not necessarily address the nuance of sign language, even simply as fingerspelling.
Therefore, dedicated training on sign language data is essential. Comparing \textit{E1.1} and \textit{E1.2}, neither is fine-tuning an existing VideoCLIP checkpoint helpful, so in the rest of the paper, models are trained from scratch.

In \textit{E2}, \textit{I3D BSL-1K} sign-language-specific features outperform \textit{HowTo100M S3D} video features (\textit{E1.2}), especially when downsampled to the same dimension as S3D (\textit{E2.1}).
\textit{MediaPipe Holistic} pose estimation as a feature extractor (\textit{E3}) works better than 3D-CNN-based video encoders, presumably because it is more universal than an I3D model pretrained on a particular dataset and sign language. 
For fingerspelling understanding, focusing on the dominant hand (\textit{E3.1}) is beneficial\footnote{Note that the DGS finger alphabet is one-handed.}, which drastically reduces the number of keypoints from 543 to 21. 
2D data augmentation further improves the overall performance.
Since \textit{MediaPipe Holistic} as features perform the best and are more interpretable and operable for potential data normalization and augmentation, we decide to use it as the frozen video encoder $f_{\theta_{ve}}$ for the rest of the paper\footnote{An S3D/I3D model fine-tuned end-to-end may overcome some limitations of pose estimation and yield superior performance for specific tasks. In this work, we trade pursuing state-of-the-art numbers for the universality, interpretability, and cheap computation of pose estimation.}.

\section{SignCLIP Pretraining}
\label{sec:sign_clip}

\begin{table*}[ht]
\centering
\resizebox{\linewidth}{!}{%
\begin{tabular}{lllrrrr}
\toprule
\textbf{Dataset} & \textbf{Language} & \textbf{Type/Task} & \textbf{\#examples} & \textbf{\#signs/\#classes} & \textbf{\#signers} \\ \midrule

\href{https://www-i6.informatik.rwth-aachen.de/aslr/fingerspelling.php}{RWTH German Fingerspelling} \citep{dreuw2006modeling} \checkmark & DGS & Isolated Fingerspelling & 1,400 & 35 & 20 \\ \midrule

\href{https://home.ttic.edu/~klivescu/ChicagoFSWild.htm}{ChicagoFSWild} \citep{shi2018american} & ASL & Continuous fingerspelling & 7,304 & -- & 160 \\ 
\href{https://home.ttic.edu/~klivescu/ChicagoFSWild.htm}{ChicagoFSWild+} \citep{shi2019fingerspelling} & ASL & Continuous fingerspelling & 55,232 & -- & 260 \\ 

\href{https://www.kaggle.com/competitions/asl-fingerspelling/data}{Google -- American Sign Language Fingerspelling Recognition} & ASL & Continuous fingerspelling & 67,208 & -- & 100 \\ 

\midrule

\href{https://dxli94.github.io/WLASL/}{WLASL} \citep{li2020word} & ASL & ISLR & 21,083 & 2,000 & 100 \\ 

\href{https://www.microsoft.com/en-us/research/project/asl-citizen/}{ASL Citizen} \citep{desai2023asl} \checkmark & ASL & ISLR & 83,399 & 2,731 & 52 \\
\href{https://github.com/leekezar/SemLex}{Sem-Lex} \citep{kezar2023sem} \checkmark & ASL & ISLR & 91,148 & 3,149 & 41 \\

\href{https://www.kaggle.com/competitions/asl-signs/data}{Google -- Isolated Sign Language Recognition (i.e., \textit{asl-signs})} \checkmark & ASL & ISLR & 94,477 & 250 & 21 \\ 
\href{https://signdata.cc.gatech.edu/view/datasets/popsign_v1_0/index.html}{Google -- PopSign ASL v1.0} \citep{starner2024popsign} \checkmark & ASL & ISLR & 200,686 & 250 & 47 \\ 

\midrule

\href{https://how2sign.github.io/}{How2Sign} \citep{Duarte_CVPR2021} & ASL & Continuous signing & 35,000 & 16,000 & 11 \\ 

\midrule

\href{https://asl-lex.org/about.html}{ASL-LEX} \citep{sehyr2021asl} & ASL & Dictionary (phonological) & 2,723 & 2,723 & (unknown) \\ 

\textbf{\href{https://www.spreadthesign.com/en.us/search/}{Spreadthesign} (SignCLIP, our filtered version)} \checkmark & Multilingual & Dictionary & 456,913 & 18,423* & (unknown) \\ 

\midrule

\href{https://www.image-net.org/}{ImageNet} \citep{deng2009imagenet} & -- & Image classification & 1,431,167 & 1,000 & -- \\ 

\href{https://www.di.ens.fr/willow/research/howto100m/}{HowTo100M} \citep{miech19howto100m} (VideoCLIP) & -- & Video understanding & 136,000,000 & -- & -- \\ 

\href{https://openai.com/index/clip/}{CLIP} \citep{radford2021learning} & -- & Contrastive learning & 400,000,000 & -- & -- \\ 

\bottomrule
\end{tabular}
}
\caption{Summarization of datasets consisting of relatively short-duration video examples, compared with Spreadthesign and common CV datasets. SignCLIP has been tested with the datasets marked with a checkmark. \textit{asl-signs} is a subset of \textit{PopSign ASL v1.0}. \textit{\#signs/\#classes} for Spreadthesign is marked with an asterisk (*) since the signs of a concept across different sign languages are barely classified as one sign.}
\label{tab:datasets}
\end{table*}

To fully realize the power of contrastive learning, 
we train and evaluate SignCLIP on larger and more diverse sign language datasets.
For efficient experimenting, we start exploring datasets consisting of relatively short-duration sign language video examples, e.g., for the task of isolated sign language recognition (ISLR) instead of machine translation.
In Table \ref{tab:datasets}, we summarize recent large-scale sign language datasets in this context, focusing on ASL, one of the highest-resourced languages in SLP.

\subsection{Spreadthesign Pretraining Dataset}
\label{subsec:sp}

Spreadthesign is used as the pretraining dataset for its large-scale and multilingual nature\footnote{\url{https://www.spreadthesign.com/en.us/about/statistics/}. The data used in this work was crawled in 2023 and might differ slightly from the official statistics.}.
In this work, we limit the text translations to English only to avoid a cartesian product number of data points, for the pretraining to be economical and sign-language-focused.
After filtering English text, our dataset consists of 18,423 concepts in 41 sign languages, resulting in 456,913 video-text pairs with a total duration of $\sim$500 hours. The data distribution is presented in Appendix \ref{appendix:sp} in detail.
We split all examples randomly into training, validation, and test sets at the ratio of 98:1:1.
A caveat of this dataset is that there is normally only one signing example per text concept per sign language, which means the pretraining is prone to overfitting the exact signs by particular signers. 
We still believe that the diverse text-signing pairs will guide the model to learn a useful visual representation.

We add the Spreadthesign data scale and that of CLIP and VideoCLIP to Table \ref{tab:datasets} for comparison. 
CLIP was trained on 400 million text-image pairs collected from the Internet (500K queries and up to 20K pairs per query); VideoCLIP was pretrained on 1.2 million videos from HowTo100M (each lasts $\sim$6.5 minutes with $\sim$110 clip-text pairs).
We additionally add ImageNet \citep{deng2009imagenet}, which has a relatively closer scale to Spreadthesign.

\subsection{Training and Evaluation Details}

Most implementation details and evaluation protocols in FingerCLIP (\S\ref{sec:finger_clip}) are reused for SignCLIP.
The text prompts now consist of the text content prepended with a spoken and a sign language tag, inspired by multilingual machine translation in \citet{Johnson2017}. For example, the text prompt for signing the phrase ``Hello, can I help you?'' in ASL is ``<en> <ase> Hello, can I help you?'', tagged by the ISO 639-3 language code.
Fitting most examples, we limit the context length of the video Transformer $f_{\theta_{v}}$ to 256 tokens, equivalent to a 10-second, 25 FPS video clip;
and that of the text Transformer $f_{\theta_{t}}$ to be 64. Both Transformers are initialized with the pretrained \textit{bert-base-cased} weights
and trained on an NVIDIA A100-SXM4-80GB GPU to maximally afford a batch size of 448\footnote{For reference, CLIP was trained with batch size 32,768 and VideoCLIP was trained with batch size 512.}.
We also measure the training efficiency by the number of parameters and the training time. 

For evaluation, we include the same video-to-text metrics used in FingerCLIP and omit text-to-video for simplicity, as the latter correlates with and is more trivial than the former. 
We additionally test the models with three recent ASL ISLR datasets in a zero-shot way to evaluate out-of-domain generalization. 
We perform the following text preprocessing to mitigate the effect of shifted text distribution when building text prompts from the raw gloss labels: (1) lowercasing the glosses; (2) removing the gloss index for different variants of a sign\footnote{This operation might not be desired if the objective is sign classification as different variants of a sign present different visual forms. The main goal here is to test how well the sign embeddings align to text semantically regardless of the form.}; (3) filtering their test sets by known text labels in Spreadthesign, which reduces the total number of signs/classes from the original test datasets.

Starting from a baseline setup that resembles \textit{E3} in FingerCLIP, we increase the video Transformer layers from 6 to 12 and try linear multimodal projection layers after temporal pooling. 
For pose data, we always simplify the face by using the contour keypoints only, resulting in 203 keypoints.
We further experiment with the following modifications: 

\paragraph{Pose Data Preprocessing}

Before the regular normalization, $D_{shoulders}=1, mid\mbox{--}point=(0, 0)$, as performed in all experiments, we further try (1) removing redundant keypoints and repositioning the wrist to the hand model's prediction (\textit{E6}); (2) standardizing the keypoints by subtracting the mean pose values of all examples from Spreadthesign and dividing by the standard deviation (\textit{E6.1}); (3) anonymizing by removing the appearance from the first frame then adding the mean pose (\textit{E6.2}).

\paragraph{Pose Data Augmentation}

After data normalization, we also employ pose-based data augmentation inspired by \citet{bohavcek2022sign} at training time to improve the models' robustness, including
(1) randomly flipping the poses horizontally; 
(2) 2D spatial augmentation as done in FingerCLIP; 
(3) temporal augmentation of the signing speed by linear interpolation between frames;
(4) Gaussian noise on keypoints.




\subsection{Experimental Results and Discussion}

\begin{table*}[htbp]

\centering
\resizebox{\textwidth}{!}{%
\begin{tabular}{ll|cccc|ccc|cc}

\toprule

 & & \multicolumn{4}{c}{\textbf{Video-to-text (In-domain)}} & \multicolumn{3}{c}{\textbf{Video-to-text (out-of-domain)}} & \multicolumn{2}{c}{\textbf{Efficiency}} \\
 
\cmidrule(lr){3-6} \cmidrule(lr){7-9} \cmidrule(lr){10-11}

\multicolumn{2}{l}{\textbf{Experiment}} & \textbf{R@1}$\uparrow$ & \textbf{R@5}$\uparrow$ & \textbf{R@10}$\uparrow$ & \textbf{MedianR}$\downarrow$ & \textbf{AS MedianR}$\downarrow$ & \textbf{AC MedianR}$\downarrow$ & \textbf{SL MedianR}$\downarrow$ & \textbf{\#Params}$\downarrow$ & \textbf{Time}$\downarrow$ \\
\midrule

\textbf{E4} & Baseline & 0.33 & 0.64 & 0.77 & 3/3939 & 103/213 & 253/1625 & 455/1967 & \textbf{175M} & 29h \\
\midrule

\multicolumn{2}{l}{\textit{Initial architectural changes}} & \\
\textbf{E5} & \textbf{E4} + six more video layers & 0.37 & 0.68 & 0.80 & \textbf{2} & 104 & 192 & 382 & 217M & 28h \\
\textbf{E5.1} & \textbf{E5} + multimodal projection layer & 0.38 & 0.69 & 0.80 & \textbf{2} & 104 & 216 & 418 & 218M & 15h  \\
\midrule

\multicolumn{2}{l}{\textit{Pose data preprocessing}} & \\

\textbf{E6} & \textbf{E5} + keypoint reduction \& reposition & 0.37 & 0.68 & 0.80 & \textbf{2} & 105 & 230 & 665 & 217M & 32h \\

\textbf{E6.1} & \textbf{E6} + keypoint standardization & \textbf{0.40} & \textbf{0.71} & \textbf{0.83} & \textbf{2} & \textbf{99} & 273 & 551 & 217M & \textbf{14h} \\

\textbf{E6.2} & \textbf{E6.1} + pose anonymization & 0.37 & 0.68 & 0.79 & \textbf{2} & 101 & 251 & 577 & 217M & 40h \\
\midrule

\multicolumn{2}{l}{\textit{Pose data augmentation}} & \\
\textbf{E7} & \textbf{E5} + pose random flipping ($p=0.2$) & 0.36 & 0.67 & 0.79 & \textbf{2} & 105 & 200 & 435 & 217M & 29h \\
\textbf{E7.1} & \textbf{E5} + spatial 2D augmentation ($\sigma=0.2$) & 0.35 & 0.65 & 0.78 & 3 & 102 & 219 & 377 & 217M & 39h \\
\textbf{E7.2} & \textbf{E5} + temporal augmentation ($\sigma=0.2$) & 0.39 & 0.69 & 0.80 & \textbf{2} & 104 & \textbf{187} & 372 & 217M & 62h \\
\textbf{E7.3} & \textbf{E5} + Gaussian noise ($\sigma=0.001$) & 0.37 & 0.68 & 0.80 & \textbf{2} & 104 & 198 & 364 & 217M & 29h \\
\midrule


\multicolumn{2}{l}{\textit{Test-time-only normalization}} & \\
\textbf{E8*} & \textbf{E7.2} + flipping to right-handed & 0.39 & 0.69 & 0.80 & \textbf{2} & 103 & \textbf{187} & \textbf{359} & -- & -- \\
\textbf{E8.1*} & \textbf{E8} + pose anonymization (zero-shot-only) & -- & -- & -- & -- & 101 & 214 & 380 & -- & -- \\

\bottomrule
\end{tabular}
}
\caption{SignCLIP experimental results evaluated on the test set. \textit{R@k} denotes \textit{recall@k}, and \textit{MedianR} denotes the median retrieval rank as well as the total number of unique signs/classes. \textit{AS} = \textit{asl-signs}, \textit{AC} = \textit{ASL Citizen}, \textit{SL} = \textit{Sem-Lex}. Experiments marked with an asterisk (*) are test-time only. The best score of each column is in bold.}
\label{tab:sign_clip}
\end{table*}

We present the experimental results in Table \ref{tab:sign_clip}. 
In this scenario, an accurate retrieval is harder than FingerCLIP because there are 3,939 unique text prompts in the test set of 4,531 examples.
The in-domain results, \textit{E6.1} at the top (attained with increased video layers and keypoint standardization), are impressive given the challenging nature. 
We attribute it to (1) the hypothesized multilingual transfer effect thanks to sign language iconicity; and (2) the broader supervision signal of contrastive learning than fixed labels, coming from example phrases consisting of individual signs (Figure \ref{fig:title}). 

On the other hand, zero-shot performance on out-of-domain data is deficient. We posit that to reach noticeable performance on out-of-domain data, few-shot learning or fine-tuning (\S\ref{subsec:islr}) is essential given the current scale of pretraining. 
Nevertheless, starting from \textit{E7}, we experiment with a few data augmentation techniques to increase data variation given that in-domain standardization benefits in-domain results but hurts out-of-domain results. We then attempt test-time-only normalization to shift the test distribution of the poses closer to training.
As a result, we manage to gradually fight against overfitting to the pretraining dataset and improve the overall zero-shot performance by temporal augmentation (\textit{E7.2}) and test-time pose flipping if the right hand is not present (\textit{E8}).
The former provides robustness to signing speed change in videos, and the latter is helpful for unseen test examples signed by left-handed signers.

\section{Downstream Tasks and Analysis}
\label{sec:evaluation}

In this section, we evaluate SignCLIP on several downstream tasks and datasets, and we discuss the ideas for further enhancement and evaluation in \S\ref{sec:limitation}.

\subsection{Isolated Sign Language Recognition}
\label{subsec:islr}

We provide a comprehensive evaluation for ASL ISLR datasets in Table \ref{tab:recognition}. 
For zero-shot prediction, we follow the optimal setups in \textit{E8}/\textit{E8.1} but skip any text preprocessing on the raw glosses for a fairer comparison.
For few-shot learning, we randomly sample 10 example videos for each class from training data and use a nonparametric k-nearest neighbors (KNN) algorithm to infer test labels based on pose embedding similarity ($n\_neighbors = \#classes$).
We additionally train a supervised logistic regression classifier with all default settings offered by \textit{scikit-learn}
on top of the embedded poses, also known as linear probe \citep{alain2016understanding}.
We also train SignCLIP models from scratch or fine-tune them from the \textit{E7.2} checkpoint with the specific downstream datasets for 25 epochs with batch size 256.
Outliers longer than 256 frames are removed.

In general, proper zero-shot prediction is hindered by distribution shift on both modalities: 
(1) for \textit{asl-signs} dataset, zero-shot prediction is flawed because we only have pose data normalized in an unknown way from Kaggle instead of the raw videos;
(2) \textit{PopSign ASL}, the superset of \textit{asl-signs} with raw videos, is released later and the evaluation on it (category \textit{game} only) is therefore superior and more meaningful;
(3) \textit{ASL Citizen} uses all upper case glosses and \textit{Sem-Lex} uses snake case glosses, both unseen for the text encoder during training.

\begin{table}[!ht]
    \centering
    \resizebox{\linewidth}{!}{%
        \begin{tabular}{ll|cccc}
        \toprule
        
        \textbf{Data} & \textbf{Model} & \textbf{R@1}$\uparrow$ & \textbf{R@5}$\uparrow$ & \textbf{R@10}$\uparrow$ & \textbf{MedianR}$\downarrow$ \\ 
        \midrule
        
        \multirow{7}{*}{\textbf{AS}} & SignCLIP (E8.1) zero-shot  & 0.01 & 0.03 & 0.05 & 118/250 \\
        & SignCLIP (E8.1) 10-shot & 0.01 & 0.03 & 0.06 & 109 \\
        & SignCLIP (E8.1) + linear  & 0.12 & 0.30 & 0.41 & 17 \\ 
        \cmidrule{2-6}
        
        & SignCLIP train from scratch & 0.74 & 0.91 & \textbf{0.94} & \textbf{1} \\
        & SignCLIP fine-tuned & 0.74 & 0.91 & \textbf{0.94} & \textbf{1} \\ 
        & SignCLIP fine-tuned + linear & \textbf{0.78} & \textbf{0.92} & \textbf{0.94} & \textbf{1}  \\ 

        \cmidrule{2-6}
        & SOTA Kaggle competition & 0.89* & -- & -- & -- \\
        \midrule

        \multirow{7}{*}{\textbf{PS}} & SignCLIP (E8) zero-shot  & 0.03 & 0.10 & 0.16 & 62/250 \\
        & SignCLIP (E8) 10-shot & 0.04 & 0.14 & 0.22 & 44  \\
        & SignCLIP (E8) + linear & 0.31 & 0.58 & 0.69 & 4 \\ 
        \cmidrule{2-6}
        
        & SignCLIP train from scratch & 0.83 & \textbf{0.97} & \textbf{0.99} & \textbf{1}  \\
        & SignCLIP fine-tuned & 0.84 & \textbf{0.97} & 0.98 & \textbf{1} \\ 
        & SignCLIP fine-tuned + linear & \textbf{0.85} & \textbf{0.97} & 0.98 & \textbf{1} \\ 
        \cmidrule{2-6}

        & SOTA PopSign ASL & 0.84 & -- & -- & -- \\

        \midrule
        \midrule
        
        \multirow{7}{*}{\textbf{AC}} & SignCLIP (E8) zero-shot & 0.00 & 0.00 & 0.00 & 1296/2731 \\
        & SignCLIP (E8) 10-shot & 0.05 & 0.16 & 0.23 & 56 \\
        & SignCLIP (E8) + linear & 0.23 & 0.47 & 0.57 & 7  \\ 
        \cmidrule{2-6}
        
        & SignCLIP train from scratch & 0.39 & 0.71 & 0.80 & 2 \\
        & SignCLIP fine-tuned & 0.46 & 0.77 & 0.84 & 2 \\ 
        & SignCLIP fine-tuned + linear & \textbf{0.60} & \textbf{0.84} & \textbf{0.89} & \textbf{1} \\

        \cmidrule{2-6}
        & SOTA ASL Citizen & 0.63 & 0.86 & 0.91 & -- \\ 

        \midrule
        \midrule
        
        \multirow{8}{*}{\textbf{SL}} & SignCLIP (E8) zero-shot & 0.01 & 0.02 & 0.03 & 853/3837 \\
        & SignCLIP (E8) 10-shot & 0.02 & 0.06 & 0.10 & 235 \\
        & SignCLIP (E8) + linear & 0.15 & 0.29 & 0.36 & 38 \\ 
        \cmidrule{2-6}
        
        & SignCLIP train from scratch & 0.14 & 0.30 & 0.38 & 32 \\
        & SignCLIP fine-tuned & 0.16 & 0.34 & 0.42 & 22 \\ 
        & SignCLIP fine-tuned + linear & \textbf{0.30} & \textbf{0.48} & \textbf{0.55} & \textbf{6} \\
        \cmidrule{2-6}

        & SOTA Sem-Lex & 0.67* & -- & -- & -- \\ 
        & SOTA + auxiliary phonology & 0.69* & -- & -- & -- \\ 

        \bottomrule
        
        \end{tabular}
    }
    \caption{Comprehensive evaluations of SignCLIP on the test set of ASL ISLR datasets. For ISLR, \textit{recall@k} equals top-n accuracy. \textit{AS} = \textit{asl-signs}, \textit{PS} = \textit{PopSign} (the superset of \textit{AS}), \textit{AC} = \textit{ASL Citizen}, \textit{SL} = \textit{Sem-Lex}. The best score SignCLIP achieves on each dataset is in bold. SOTA numbers marked with an asterisk (*) are not directly comparable to ours. \textit{SOTA Kaggle} is trained with \textit{asl-signs} but tested on a private test set; \textit{SOTA Sem-Lex} is tested with a reduced test set of 2,731 classes that are aligned with \textit{ASL Citizen}/\textit{ASL-LEX} and is thus considered an easier objective.}
    \label{tab:recognition}
\end{table}

For \textit{asl-signs}, we find pose anonymization eases the domain shift issue in zero-shot (\textit{E8.1}) and other settings.
Besides, pose-based few-shot KNN greatly improves the deficient zero-shot results, bypassing the influence of out-of-domain glosses.
Finally, fine-tuning SignCLIP with a pretrained checkpoint, compared to training from scratch on the target dataset, yields more competitive or even better results than the state-of-the-art (SOTA) reported in the previous literature.

\subsection{Sign Language Identification}

Sign language identification \citep{gebre2013automatic} can be achieved by simply ranking text prompts of different sign languages without actual content, e.g., ``<en> <ase>'' for ASL.
We evaluate the best checkpoint \textit{E6.1} for in-domain test data and obtain 0.99 \textit{recall@1}, which solves the task perfectly without any direct supervision.
However, the identification task is ill-defined in that the model can learn to identify signers for particular sign languages.
This task is considered a simplified version of in-domain video-to-text retrieval, as it eliminates the need to simultaneously distinguish language and concept.

















\subsection{Latent Space Exploration}

We make SignCLIP accessible via a RESTful API and perform analysis in a Colab notebook\footnote{\url{https://colab.research.google.com/drive/1r8GtyZOJoy_tSu62tvi7Zi2ogxcqlcsz?usp=sharing}}.

\paragraph{Distributional Hypothesis for Sign Language}

\begin{figure}
\centering
\includegraphics[width=\linewidth]{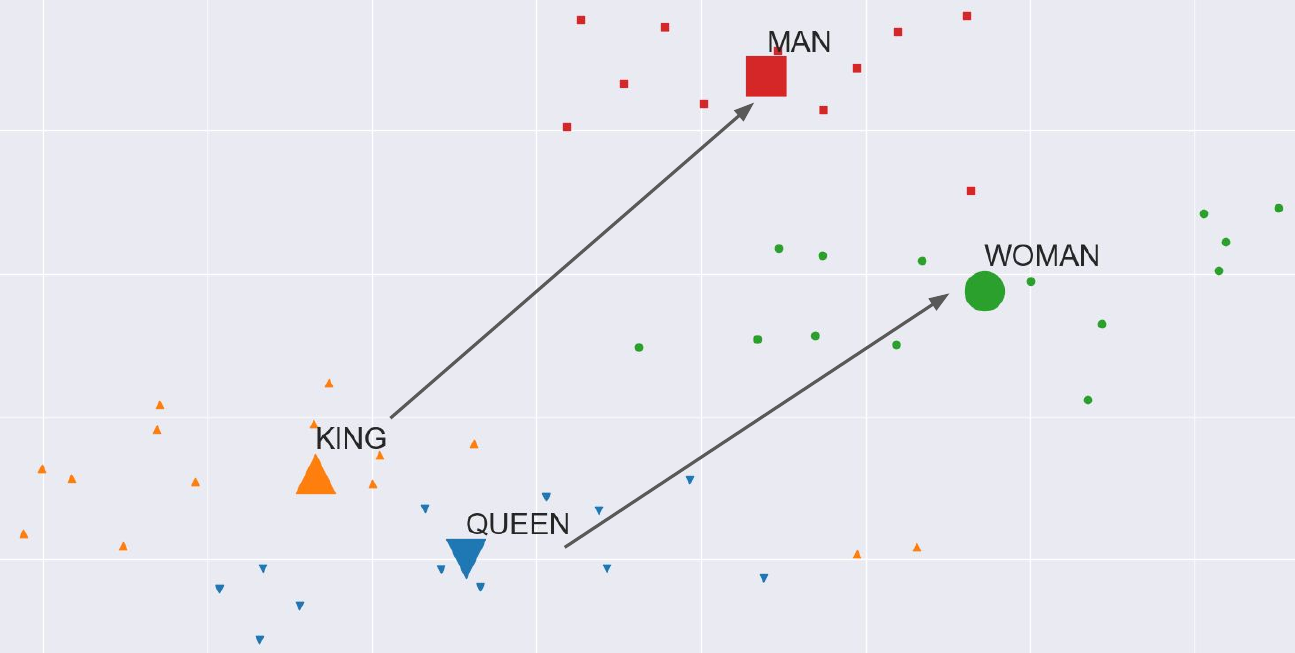}
\caption{\textit{King – Man + Woman = Queen} analogy revisited. 14 video examples of each sign are randomly sampled from the ASL Citizen dataset, embedded by a fine-tuned SignCLIP pose encoder, and then visualized by \textit{t-SNE (perplexity=15)} with different shapes and colors. Cluster centers are represented with a big symbol.} 
\label{fig:latent}
\end{figure}

We revisit the distributional hypothesis \citep{lenci2023distributional} in a sign language context based on the pose/sign embeddings instead of text/word \citep{mikolov2013distributed}.
As illustrated by Figure \ref{fig:latent}, unlike a word with a discrete, unique text token, each sign has multiple realizations scattered in a continuous space.
By aligning pose representation to text with contrastive learning, we have realized distributional semantics in sign language, reflected by the cluster center of each sign, while maintaining individual variance. 

\paragraph{What Is the Most Iconic Sign Crosslingually?}

Iconicity is one of the key motivations for training a multilingual SignCLIP (Figure \ref{fig:title}), and now we ask this linguistic question back to the model. 
We rank a sampled subset of 302 signs from \textit{Spreadthesign} based on the variance of the pose embeddings across 20+ different sign languages.
As a result, the sign for ``scorpion'' with a universal hook hand shape ranks at the top, and the motivational example, i.e., the ``house'' sign, also ranks high (51/302).
Conversely, the signs for numbers rank low due to the diverse signing styles across languages.
We append the full rank in Appendix \ref{appendix:rank}.

\section{Related Work}

Contrastive learning has been increasingly used in contemporary SLP work.
We discuss some related work in this session, which aims mainly at improving performance on dedicated tasks and datasets. Therefore, their goals are slightly different than SignCLIP, a universal pretrained sign language embedding model aligned to spoken language text.

\citet{gan2023contrastive,zheng2023cvt,ye2024improving} focus on the small-scale RWTH-PHOENIX-Weather 2014T and CSL-Daily datasets.
\citet{gan2023contrastive} designs a visual contrastive loss and a semantic contrastive loss to tackle the CTC spike
phenomenon in the Continuous Sign Language Recognition (CSLR) task and the exposure bias issue in the Sign Language
Translation (SLT) task.
\citet{zheng2023cvt} proposes an explicit contrastive cross-modal alignment between video frames and glosses in combination with an implicit cross-modal alignment by a variational autoencoder (VAE) for CSLR.
\citet{ye2024improving} identifies a visual representation density issue in SLT and introduces a frame-wise contrastive learning strategy to alleviate the issue and improve SLT. 

\citet{wong2023learnt} has a similar sign embedding idea as SignCLIP, namely \textit{Learnt Contrastive Concept}, that aligns with word embeddings of the linguistic labels for sign video and incorporates this into ISLR pipelines for the WLASL and BOBSL datasets.
\citet{raude2024}, on the other hand, tackles CSLR on BOBSL by a multi-task setting, including two contrastive losses for sign-level and sentence-level retrieval, respectively. They show that joint training for CSLR and sign language retrieval is mutually beneficial.

Like our pretraining task framing introduced in \S\ref{sec:finger_clip}, \citet{cheng2023cico} explicitly formulates text-to-video/video-to-text retrieval as a cross-lingual contrastive learning task.
They address the data scarcity issue by combining a domain-agnostic and a domain-aware video encoder and show its effectiveness on How2Sign and PHOENIX-2014T.




\section{Conclusion: Where Are We for SLP?}

This work involves SLP, an interdisciplinary field that suffers from low-resourceness compared to mainstream NLP and CV. 
To overcome the non-generalizability of a specific dataset/task/language, we adapt (Video-)CLIP and propose SignCLIP.
SignCLIP is trained on \textit{Spreadthesign}, a multilingual, generic sign language dictionary consisting of $\sim$500 hours of signing in 41 sign languages, and is evaluated extensively for various purposes.

SignCLIP demonstrates excellent in-domain performance but falls short of immediate zero-shot prediction on downstream ISLR tasks. 
This finding is consistent with previous CV studies \citep{li2017learning} before CLIP reached its scale.
Similar models trained on smaller datasets close to an ImageNet scale performed much worse than supervised baselines on common benchmarks.
By comprehensively evaluating downstream ASL ISLR performance (Table \ref{tab:recognition}), we also intend to shed some light on data efficiency, as a fully supervised approach usually requires meticulous, task-specific data collection.
This is more demanding for SLP, a niche domain lacking human experts.

As a middle ground between full zero-shot prediction and full supervision, few-shot learning or fine-tuning is essential to tackle domain shift and is more realistic given the present methodology and data scale.
The cross-lingual transfer effect is prospective for sign language thanks to iconicity.
With no surprise, under a sensible data condition, the universal pretraining paradigm that is transforming NLP and CV is also a promising research direction for SLP.



\section{Limitations}
\label{sec:limitation}

The main limitation of this work is the data scale, which interestingly is the primary breakthrough of the original CLIP work.
Apart from the inherent data scarcity issue in SLP research, the concrete limitations of SignCLIP come in a few aspects.

First, to be not dataset/task/language-specific while possibly large-scale, we choose \textit{Spreadthesign} as our pretraining dataset (\S\ref{subsec:sp}), which is indeed highly multilingual and untied to any SLP task. 
However, we cannot release the dataset publicly, and future researchers who are interested in it have to first resolve or purchase the license from \textit{Spreadthesign}.
Fortunately, it is possible to augment or replace \textit{Spreadthesign} with recently released public datasets of comparable or larger size, which should increase data density and variation sign-wise, but with potentially decreased diversity and balance among different sign languages.
Although pretrained with a highly multilingual dataset, this paper focuses downstream task evaluation on ASL, one of the highest-recourse sign languages. We leave further evaluation and exploration of other sign languages to future work.

Secondly, training large models on video data is very costly. The principle of this work is to finish every training process in less than three days on a single Nvidia A100 GPU, and we managed this goal (Table \ref{tab:sign_clip}) by using pose-based models, a moderate context length of 256 frames, and limiting spoken language to English only.
An ablation study on video length would be interesting to help understand the limitations of the current static embedding approach\footnote{This is also relevant to the polysemy phenomenon between signs and words, i.e., one sign can mean more than one word and one word can also correspond to more than one sign.}.
To scale up, multiple GPU training can be exploited, and architectural modifications that reduce the sequence length must be employed, for which we refer to several techniques discussed in \S\ref{sec:background}.

As a result of the above-mentioned optimizations, we will be able to remove the limitation of the relatively short context, and then train and evaluate for longer-range tasks such as machine translation and sign language production.
This will make SignCLIP more versatile and generalizable, as well as support other types of deep pretrained networks for SLP, such as large language models.

\section*{Acknowledgments}

This work is funded by the Swiss Innovation Agency  (Innosuisse) flagship IICT (PFFS-21-47) and by the SIGMA project at the UZH Digital Society Initiative (DSI).

We thank the (meta-)reviewers for their valuable feedback. We also thank Colin Leong, Gomèr Otterspeer, Jetske Adams, Oline Ranum, and Hamzah Luqman for their immediate interest and lively discussion in our work.

\bibliography{acl_latex}

\appendix

\onecolumn

\section{An Intuitive Test on ChatGPT's Sign Language Understanding Ability}
\label{appendix:chatgpt}

\begin{figure}[ht]
    \centering
    \includegraphics[width=\linewidth]{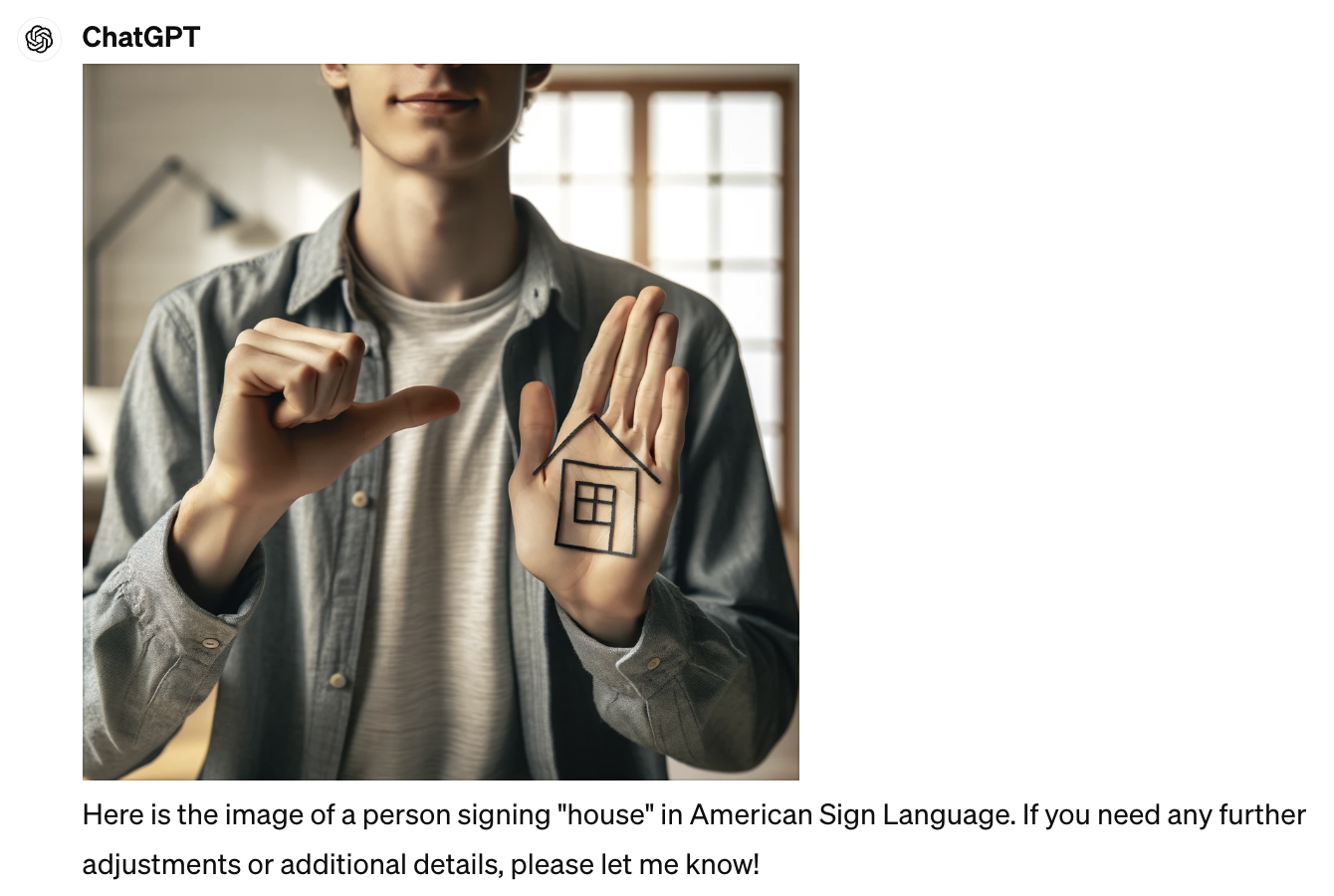}
    \caption{Screenshot of prompting ChatGPT 4o to sign ``house'' in ASL, which lacks sign language knowledge and tries to sketch a picture of a house on the open palm, tested in June 2024.}
    \label{fig:house}
\end{figure}

\newpage

\section{Introduction of the RWTH German Fingerspelling Database}
\label{appendix:rwthfs}

RWTH German Fingerspelling Database contains $\sim$1400 videos of 20 different signers. 
Each video was recorded by one webcam named \textit{cam1}, recording the dominant hands only
with a resolution of 320x240 at 25 FPS; 
and one camcorder named \textit{cam2}, recording the whole body
with a resolution of 352x288 at 25 FPS.
We exclude all \textit{cam1} videos for pose-based models since we assume that the pose estimation system expects whole-body input.

\begin{figure}[ht]
\centering
\includegraphics[width=\linewidth]{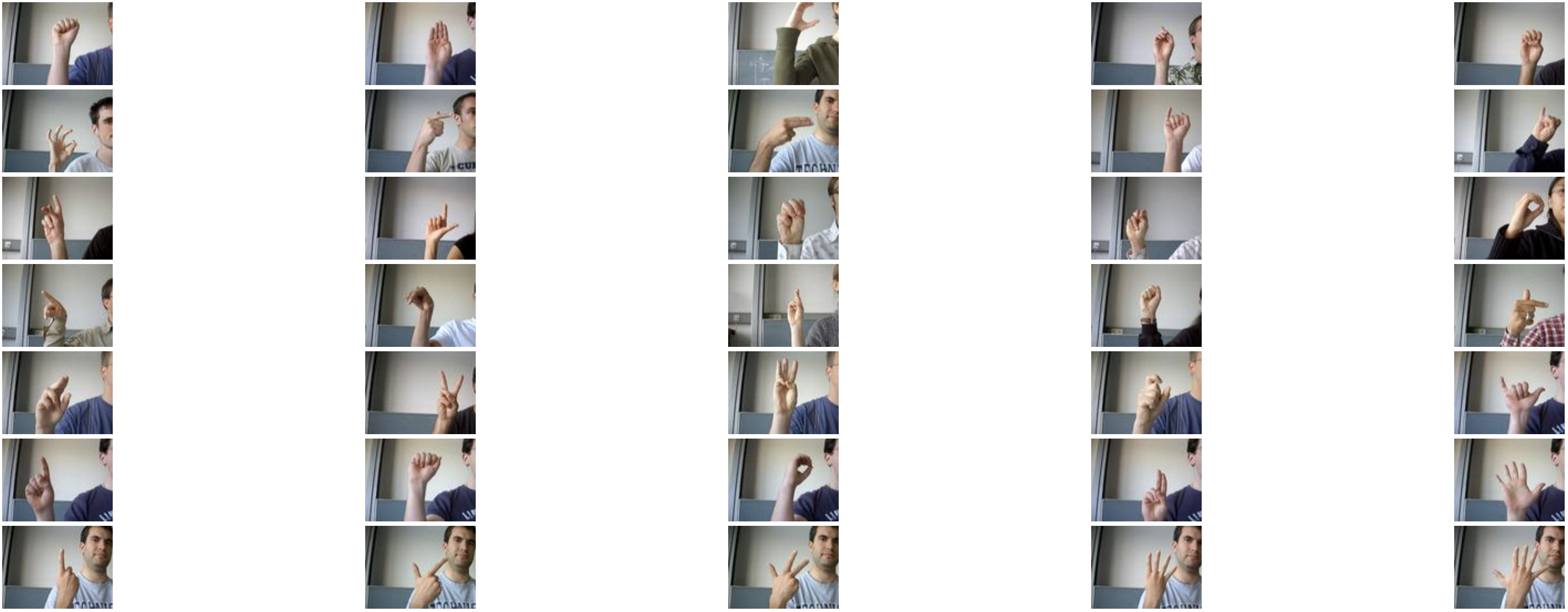}
\caption{Examples of the German finger-alphabet taken from the RWTH gesture database recorded with the webcam showing the letters A-Z, Ä, Ö, Ü, SCH, and the numbers 1 to 5. Note that J, Z, Ä, Ö, and Ü are dynamic gestures. Figure taken from \url{https://www-i6.informatik.rwth-aachen.de/aslr/fingerspelling.php}.} 
\label{fig:RWTHFS}
\end{figure}

\newpage

\section{Extended Spreadthesign Data Analysis}
\label{appendix:sp}

\begin{figure*}[!htbp]
    \centering
    \includegraphics[width=\textwidth]{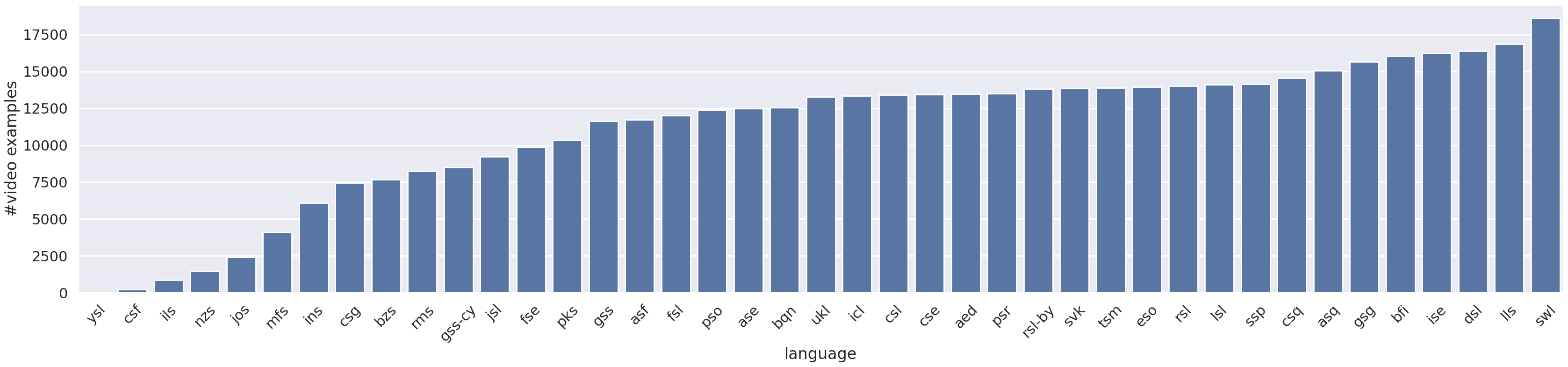}
    \caption{Sign language distribution of video examples in Spreadthesign, using the ISO 639-3 language codes.}
    \label{fig:language_distribution}
\end{figure*}

Following the data statistics presented in \S\ref{subsec:sp}, Figure \ref{fig:language_distribution} illustrates the distribution of the video examples in 41 sign languages we use from Spreadthesign.

\begin{figure*}[!htbp]
    \centering
    \includegraphics[width=\textwidth]{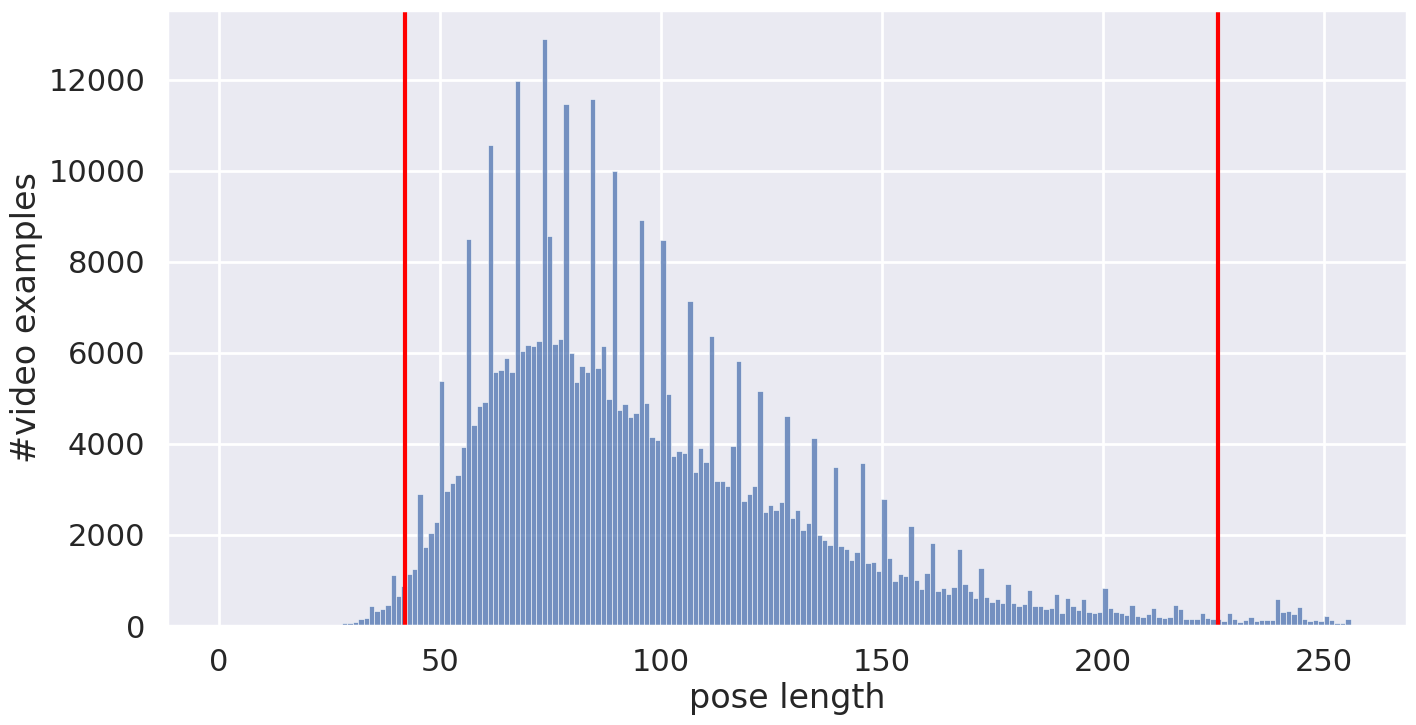}
    \caption{Pose length distribution of video examples in Spreadthesign. The two red vertical lines denote the 1st and 99th percentile of the number of frames.}
    \label{fig:pose_distribution}
\end{figure*}

Figure \ref{fig:pose_distribution} illustrates the distribution of pose/video length in Spreadthesign, depending on which we decide the pretraining context length to be 256.

\begin{figure*}[!htbp]
    \centering
    \includegraphics[width=\textwidth]{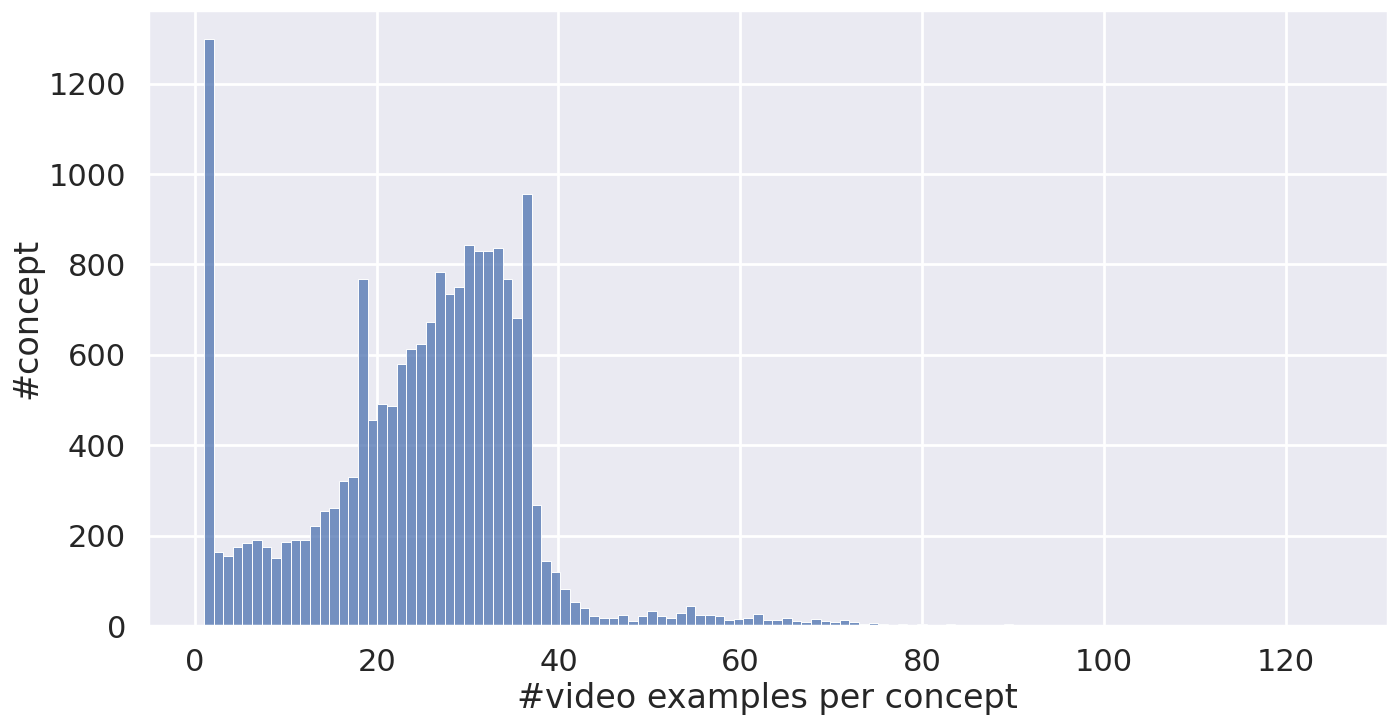}
    \caption{Concept distribution of video examples in Spreadthesign.}
    \label{fig:concept_distribution}
\end{figure*}

Figure \ref{fig:concept_distribution} illustrates the distribution of the number of video examples for 18,423 cross-lingual concepts in Spreadthesign. Most concept has only one video example, or one video example per sign language, and very few concepts have more than one video example for one specific sign language.

\clearpage
\newpage

\section{Full Rank of the Signs for Iconicity Study}
\label{appendix:rank}

\begin{lstlisting}[language=Python, basicstyle=\small,numbers=left,xleftmargin=3em,frame=single,framexleftmargin=3em]
('scorpion', 180.44417)
('reduction', 181.40424)
('illustration', 182.58504)
('envelope', 182.63124)
('observe', 183.66785)
('gold', 184.15184)
('telephone', 184.36679)
('email', 185.0535)
('center', 185.35873)
('knife', 185.70123)
('fake', 185.7079)
('frozen', 186.34874)
('high', 186.40546)
('consume', 186.42805)
('crocodile', 186.82196)
('toothbrush', 186.83427)
('Youtube', 186.87538)
('earache', 186.9703)
('Mexico', 186.98878)
('ear', 187.11588)
('Remind', 187.29926)
('Notepad', 187.38866)
('Put', 187.41467)
('potato', 187.4263)
('conceit', 187.43402)
('thong', 187.492)
('sauce', 187.50092)
('obsessed', 187.57285)
('drum', 187.8526)
('Cuba', 188.00418)
('generation', 188.16415)
('grief', 188.4431)
('guillotine', 188.59848)
('to', 188.62285)
('bind', 188.63608)
('umbrella', 188.81358)
('omit', 189.20493)
('Superman', 189.20782)
('advice', 189.48647)
('Refuse', 189.52502)
('speed', 189.6827)
('diamond', 189.76154)
('cute', 189.82486)
('headache', 190.24924)
('jealousy', 191.17343)
('flag', 191.18753)
('banana', 191.20346)
('Wait!', 191.36217)
('yet', 191.75407)
('theft', 191.75734)
('house', 191.79712)
('percentage', 191.80008)
('eye', 191.84584)
('understanding', 191.95715)
('badly', 192.02959)
('skin', 192.04294)
('dvd', 192.05695)
('until', 192.08847)
('Denmark', 192.28465)
('flower', 192.2852)
('sew', 192.4188)
('arrest', 192.55325)
('previous', 192.55621)
('neighbor', 192.59973)
('spoon', 192.78313)
('bell', 192.82666)
('click', 192.86957)
('vaccination', 192.90399)
('leading', 193.04103)
('total', 193.0793)
('internet', 193.26022)
('sandwich', 193.27394)
('erect', 193.31744)
('signature', 193.40796)
('six', 193.43289)
('strange', 193.51538)
('boy', 193.53194)
('Farewell!', 193.86534)
('cafe', 193.86969)
('twice', 193.91452)
('four', 193.96509)
('silver', 193.97311)
('China', 194.03874)
('certify', 194.06538)
('soup', 194.08856)
('rape', 194.09244)
('necessary', 194.1475)
('curious', 194.27072)
('tall', 194.3718)
('battle', 194.45622)
('promise', 194.51105)
('deadline', 194.61055)
('certificate', 194.736)
('genuine', 194.79256)
('blood', 194.82959)
('ban', 194.84705)
('uncle', 194.85689)
('quarantine', 194.90247)
('salad', 195.1925)
('ant', 195.31888)
('thief', 195.48566)
('fall', 195.60696)
('childlike', 195.6437)
('Japan', 195.68546)
('run', 195.70013)
('booking', 195.75223)
('homesick', 195.86086)
('advanced', 195.86685)
('Where?', 195.92314)
('bridge', 195.98723)
('beside', 196.01958)
('cup', 196.03467)
('spaghetti', 196.04333)
('dizziness', 196.06311)
('mixer', 196.10735)
('Assessment', 196.14081)
('amnesia', 196.15091)
('gigantic', 196.21085)
('priest', 196.2539)
('sorry', 196.31093)
('investment', 196.35571)
('believe', 196.41522)
('hang', 196.4303)
('three', 196.43767)
('hearing', 196.51305)
('principal', 196.63132)
('punctual', 196.70624)
('adult', 196.91183)
('thin', 196.98502)
('word', 197.0534)
('arm', 197.11966)
('censorship', 197.13493)
('several', 197.31216)
('bewilder', 197.43254)
('reply', 197.46024)
('serious', 197.57898)
('sewing', 197.62848)
('do', 197.69405)
('together', 197.83093)
('hairgrowth', 197.85425)
('bull', 197.98575)
('honeymoon', 198.02545)
('ball', 198.05637)
('ancient', 198.14444)
('selfish', 198.15906)
('arise', 198.17198)
('wedding', 198.21584)
('hour', 198.27957)
('granddaughter', 198.29315)
('circle', 198.32275)
('couch', 198.38385)
('scientist', 198.44269)
('important', 198.52599)
('helicopter', 198.61218)
('born', 198.69475)
('trousers', 198.7119)
('acceptable', 198.82736)
('lamp', 198.86002)
('appetite', 198.86995)
('association', 198.87001)
('leave', 198.95593)
('dyslexia', 198.98013)
('twin', 199.01114)
('force', 199.04541)
('insist', 199.06236)
('vase', 199.08466)
('easter', 199.23846)
('plate', 199.27231)
('best', 199.32513)
('heal', 199.61261)
('petrol', 199.672)
('cleaner', 199.69077)
('pepper', 199.92517)
('economic', 199.97253)
('yoghurt', 200.06439)
('brother', 200.0689)
('unpleasant', 200.13562)
('grapes', 200.14981)
('buy', 200.29669)
('2', 200.31093)
('frog', 200.41377)
('committee', 200.56615)
('complain', 200.57298)
('40', 200.58508)
('faultless', 200.59167)
('letter', 200.63252)
('angel', 200.65413)
('corruption', 200.66101)
('director', 200.67601)
('export', 200.99376)
('acne', 201.08725)
('participate', 201.14157)
('injury', 201.19518)
('offline', 201.215)
('hurt', 201.2529)
('shy', 201.31795)
('kilometer', 201.32117)
('inauguration', 201.33997)
('tale', 201.4152)
('very', 201.45908)
('law', 201.47714)
('diploma', 201.56801)
('music', 201.57074)
('war', 201.63304)
('school', 201.65495)
('horse', 201.746)
('heartburn', 201.86896)
('21', 201.90666)
('surname', 201.94667)
('addicted', 201.98883)
('supper', 202.08438)
('fun', 202.09558)
('terrorist', 202.22504)
('nanny', 202.32727)
('departure', 202.34787)
('600', 202.38922)
('pet', 202.46806)
('thousand', 202.5422)
('ice', 202.56726)
('menu', 202.57507)
('revise', 202.69331)
('haired', 202.70969)
('feeling', 202.82637)
('divorced', 202.85403)
('person', 203.0279)
('dawn', 203.36467)
('anxious', 203.46112)
('autism', 203.48038)
('discussion', 203.56613)
('adoption', 203.58824)
('truth', 203.6054)
('enemy', 203.6127)
('midnight', 203.63316)
('psychology', 203.70671)
('possible', 203.8542)
('pale', 203.85995)
('cucumber', 203.87328)
('favourite', 203.95981)
('rice', 204.09598)
('bedroom', 204.11554)
('sea', 204.18881)
('shock', 204.216)
('Admitted', 204.22227)
('anxiety', 204.22739)
('ten', 204.28094)
('international', 204.30515)
('curly', 204.32364)
('alarm', 204.42891)
('corn', 204.57687)
('upset', 204.593)
('morning', 204.65657)
('spinach', 204.6566)
('celebrate', 204.74715)
('confused', 204.82571)
('25', 204.87206)
('achievement', 204.94815)
('climate', 205.02107)
('communication', 205.26228)
('delegate', 205.26901)
('doorbell', 205.41678)
('anaesthesia', 205.66046)
('world', 205.86342)
('television', 206.1195)
('information', 206.20271)
('style', 206.28078)
('chip', 206.31616)
('anonymous', 206.41803)
('fashioned', 206.46509)
('development', 206.46605)
('scarf', 206.58081)
('uploading', 206.93182)
('900', 206.93893)
('500', 207.1253)
('camera', 207.15207)
('homeless', 207.25655)
('automatic', 207.29578)
('1000000', 207.40567)
('chef', 207.72531)
('50', 207.73314)
('infant', 207.88846)
('actress', 207.97646)
('nurse', 208.75182)
('800', 208.8017)
('slow', 208.93741)
('clinic', 208.93753)
('apartment', 209.07938)
('employed', 209.48071)
('electrician', 209.54414)
('painter', 209.57893)
('desert', 209.70918)
('Audiologist', 209.97559)
('engine', 210.53745)
('barber', 210.76971)
('bathroom', 210.77377)
('diabetic', 211.66092)
('7', 211.76964)
('27', 211.83792)
('depressed', 212.88554)
('employee', 213.17842)
('8', 213.59476)
('farmer', 216.97792)
('29', 217.70363)
\end{lstlisting}

\end{document}